\documentclass[a4paper,10pt]{article}
\usepackage[top=20mm, bottom=20mm, left=22mm, right=22mm]{geometry}
\usepackage[utf8]{inputenc}
\usepackage[T1]{fontenc}
\usepackage{microtype}
\usepackage[italian]{babel}
\usepackage{amsmath}
\usepackage{amsthm}
\usepackage{algorithmic} 
\usepackage{algorithm}
\usepackage{graphicx}
\usepackage{booktabs}
\usepackage{url}
\usepackage{subfigure}
\usepackage{epstopdf}

\newtheorem{thm}{Teorema}

\makeatletter
\renewcommand*{\@fnsymbol}[1]{\ensuremath{\ifcase#1\or \dagger\or \ddagger\or
		\mathsection\or \mathparagraph\or \|\or **\or \dagger\dagger
		\or \ddagger\ddagger \else\@ctrerr\fi}}
\makeatother

\title{Viral Search: un algoritmo genetico}
\author{Matteo Gardini\thanks{matteo.gardini@mail.polimi.it.}}
\date{\today}
\begin{document}

\maketitle

\begin{abstract}
\noindent L'articolo, dopo una breve introduzione sugli algoritmi genetici e sul loro funzionamento, presenta un tipo di algoritmo genetico chiamato \textit{Viral Search}. Ne vengono presentati i concetti chiave, viene derivato formalmente l'algoritmo e sono eseguiti test numerici volti a illustrarne potenzialità e limiti.
\end{abstract}

\section{Introduzione}
Un algoritmo genetico è un algoritmo euristico ispirato alla teoria darwinaiana dell'evoluzione ed ai concetti di selezione 
naturale ed evoluzione biologica. Gli algoritmi genetici permettono di valutare le soluzioni di partenza, ricombinarle ed, introducendo elementi di disordine, produrne di nuove convergendo, così, alla soluzione ottima. \\
Nel corso degli ultimi decenni molti problemi concreti sono stati affrontati come problemi di massimizzazione o minimizzazione di una certa funzione obiettivo. Basti pensare ai vari problemi di tipo industriale di minimizzazione dei costi, di massimizzazione dei ricavi, di gestione ottima delle risorse e così via. Numerosi metodi sono stati sviluppati per risolvere il problema di trovare il massimo o il minimo di una data funzione quando essa è soggetta a vincoli. Tra essi particolarmente diffusi sono il metodo del simplesso \cite{Dantzig1963}, il metodo dei moltiplicatori di Lagrange, i metodi del gradiente \cite{Hestens1952} e del punto interno \cite{Dantzig2003}. Come è ben noto gli aspetti più critici dei problemi di ottimizzazione insorgono quando si presentano non linearità o nella funzione considerata o nei vincoli che rendono il problema non convesso, quando sono presenti più minimi o massimi locali o quando la funzione non è differenziabile: in questo ultimo caso il metodi basati sul gradiente falliscono poichè le derivate direzionali non possono essere calcolate.\\
L'uso degli algoritmi genetici risulta particolarmente adatto in tutte queste situazioni poichè essi presentano certe caratteristiche che gli permettono di aggirare i problemi sopra elelencati. In particolare essi:
\begin{itemize}
 \item permettono di risolvere un problema di ottimizzazione per il quale non è noto un efficiente metodo analitico che possa trovare una soluzione in tempo polinomiale o lineare.
 \item permettono di trovare soluzioni globalmente ottime anche se la funzione obiettivo presenta numerosi minimi locali.
 \item hanno capacità di adattamento automatico nel caso il problema sottostante cambi nel tempo.
 \item sono ideali nel caso in cui si voglia trovare il minimo di una funzione non differenziabile, poichè non necessitano del calcolo delle derivate direzionali.
 \item sono facilmente estendibili nel caso in cui si voglia affrontare un problema multiobiettivo.
 \item sono intrinsicamente parallilizzabili.
\end{itemize}

E' stata sopratutto la loro grande adattabilità e versatilità a renderli così popolari ed utilizzati. La nascita degli algoritmi genetetici trova la sua collocazione negli anni settanta quando l'informatico Ingo Rachemberg parlò per la prima volta di \textquotedblleft stategie evoluzionistiche\textquotedblright\ \cite{Rachemberg1971}. Tuttavia, per vedere la prima implementazione di un algoritmo genetico bisogna attendere il 1975 quando John Henry Holland nel suo libro \textit{Adaptation in Natural and Artificial Systems} \cite{Holland1975} presentò una serie di teorie e tecniche tuttora fondamentali. E' proprio ad Holland che si deve il così detto \textit{Teorema degli Schemi}, teorema fondamentale che assicura la convergenza degli algoritmi genetici. Pensato e dimostrato da Holland per la codifica binaria, grazie ai contributi di Wright \cite{Wright1991} ha trovato formulazione e dimostrazione anche in alcuni casi di codifica con numeri reali. Negli anni a venire gli algoritmi genetici hanno permesso un notevole sviluppo nell'intelligenza artificiale ed hanno trovato applicazione in svariati campi dalla biologia all'economia, dall'analisi numerica alla finanza quantitativa \cite{Manicardi2013}.\\ 
L'obiettivo di questo articolo è riassumere gli aspetti fondamentali degli algoritmi genetici per poi presentarne una variante che risulta essere particolarmente adatta ai computer con bassa potenza di calcolo e che permette di trovare il minimo assoluto di funzioni che presentano numerosi minimi locali o tempo varianti. Presenteremo formalmente questo algoritmo, che chiameremo \textit{Viral Search}, e lo testeremo su vari tipi funzioni generalmente usate nelle fasi di test degli algoritmi di ottimizzazione. 

\section{Funzionamento}
Come già accennato in precedenza, gli algoritmi genetici traggono ispirazione dall'evoluzione e dalla selezione naturale. Ecco perchè, molti termini tipici di tali algoritmi trovano un corrispettivo nella biologia e nella genetica. Per maggiore chiarezza espositiva li elenchiamo brevemente di seguito.

\begin{itemize}
 \item \textit{Cromosoma}: Ciascuna delle soluzioni del problema in esame: di solito è costituita da un vettore di bit o numeri reali.
 \item \textit{Popolazione}: L'insieme di soluzioni del problema considerato.
 \item \textit{Gene}: Parte di un comosoma che solitamente consiste in un elemento del cromosoma o in una parte di esso.
 \item \textit{Fitness}: Grado di valutazione associato ad una soluzione. La \textquotedblleft bontà\textquotedblright\ della soluzione 
 è valutata tramite un'apposita funzione (la funzione obiettivo) che varia a seconda del problema considerato.
 \item \textit{Crossover}: generazione di un nuovo cromosoma ottenuta mischiando due cromosomi già esistenti.
 \item \textit{Mutazione}: alterazione casuale di uno o più geni del cromosoma.
\end{itemize}

Ciò che accomuna gran parte degli algoritmi genetici è il procedimento che essi adottano per evolvere verso una soluzione ottima del problema.
Essi seguono i seguono, generalmente, i seguenti step.

\begin{itemize}
 \item Generazione casuale della popolazione di partenza. In questa prima fase viene generata casualmente una popolazione iniziale che evolverà verso la soluzione ottima di generazione in generazione.
 \item Applicazione della soluzione fitness alla popolazione. In questa fase la funzione fitness viene applicata a tutti gli individui,
 cioè ai cromosomi, della popolazione al fine di individuare gli elementi migliori.
 \item Crossover. Sulla base delle soluzioni scelte al punto precedente vengono generate soluzioni ibride.
 \item Generazione di una nuova popolazione. Viene generata una nuova popolazione sulla base delle scelte effettuate ai punti precedenti.
 \item Iterazione della procedura a partire dal secondo punto fino ad un prestabilito istante di tempo oppure fino a che non siano presenti evidenti miglioramenti nella funzione fitness.
\end{itemize}

E' probabile che un procedimento come quello appena descritto possa portare l'algoritmo a ricadere in minimi locali. Onde
evitare questa evenienza è spesso utile mutare casualmente uno o più geni del cromosoma così da evitare che le soluzioni ristagnino in
minimi locali: introducendo tali elementi di casualità l'algoritmo è in grado di esplorare altre possibili soluzioni che, potenzialmente, potrebbero essere migliori di quelle trovate fino ad ora. Generalmente il verificarsi di una mutazione è dettato da un parametro apposito che modellizza opportunamente la probabilità di mutazione. Un ulteriore accorgimento per evitare che l'algoritmo stagni in minimi locali è la tecnica dell'elitarismo: nel momento in cui viene generata la nuova popolazione, tra i nuovi individui vengono copiati anche gli individui migliori della popolazione precedente. La discussione su ognuno degli step seguiti da un algoritmo genetico classico è complessa e vasta: per dettagli ulteriori si può fare riferimento a \cite{Poli2008}, \cite{Fogel1995}, \cite{Koza1992}.
\par E' fondamentale, a questo punto, garantire la convergenza di tali algoritmi verso la soluzione ottima. Come già accennato, il 
\textit{Teorema degli schemi} di Holland, ne dimostra la convergenza.

\subsection{Il teorema degli schemi}
In questa sezione esporremo il teorema degli schemi, risultato fondamentale che prova la convergenza degli algoritmi genetici. Una discussione
completa può essere trovata in \cite{Bridges1987}. Per poter provare tale risultato è necessario definire cosa sia uno schema. Supponiamo, per semplicità, di considerare la codifica binaria e con $\star$
una generica componente di un vettore. Uno schema è costituito dall'alfabeto dei segni (in caso di codifica binaria $0$ e $1$) aumentato
di $\star$. Uno schema rappresenta quindi tutte le stringhe di uno spazio che coincidono con lo schema eccetto il carattere $\star$, ad esempio
$\left(0,1,\star\right)$ coincide con $\left(0,1,0\right)$ e con $\left(0,1,1\right)$ nell'alfabeto dei segni $\left\{0,1\right\}$.
\par Quanto trattato il questo paragrafo si applica per l'analisi di un classico algoritmo genetico con ciclo interno del seguente tipo

\begin{algorithm}
\caption{Algoritmo Genetico Classico}
\begin{algorithmic}[1]
\STATE $t=t+1$
\STATE Select new population $P\left(t\right)$ from the old one $P\left(t-1\right)$ by single string selection based on a positive fitness function. 
\STATE Ricombine $P\left(t\right)$ by crossover and mutation.
\STATE Evaluate $P\left(t\right)$.
\STATE Repeat.
\end{algorithmic}
\end{algorithm}
Definito cosa sia un algoritmo genetico classico, passiamo ora a dare alcune definizioni necessarie per comprendere il teorema.
Definiamo $\delta \left( S \right) $ la lunghezza di uno schema, cioè la distanza tra il primo e l'ultimo componente fisso. Ad esempio
$\delta \left( \left( \star, \star, 0, 1,\star ,0,1\right)\right)=5$. Sia inoltre $\sigma\left( S \right)$ l'ordine dello schema, cioè
il numero di componenti fisse: ad esempio $\sigma\left(\star, 0,1\right)=2$. Sia poi $eval\left(S,t\right)$ il fitness di uno schema,
cioè il fitness medio di tutte le stringhe della popolazione che contengono lo schema $S$.  Ad esempio se lo schema $S$ contiene $n$ individui
$\left\{v_1,\dots,v_n\right\}$ allora
\begin{equation}
 eval(S,t)=\frac{1}{n} \sum_{k=1}^{n} eval\left(v_k\right),
\nonumber
\end{equation}
dove $eval\left(v_k\right)$ è il fitness dell'individuo $k$. Siano inoltre $N\left(t\right)$ la dimensione della popolazione in $t$ e 
$F\left(t\right)=\sum_{k=1}^n eval\left(v_k\right)$ il fitness totale della popolazione.\\
Possiamo dedurre allora che ogni individuo hai probabilità $eval\left(v_k\right)/F\left(t\right)$ di essere selezionato. Definito allora
$\xi\left(S,t\right)$ il numero di individui all'istante $t$ che presentano lo schema $S$ ci si può aspettare che
\begin{equation}
 \xi\left(S,t+1\right)=\xi\left(S,t\right)N\left(t\right) \frac{eval\left(S,t\right)}{F\left(t\right)}.
\label{Eq_CrescitaSchema}
\end{equation}
Se definiamo il fitness medio della popolazione come $\bar{F}\left(t\right)=\frac{F\left(t\right)}{N\left(t\right)}$ 
l'Equazione $\eqref{Eq_CrescitaSchema}$ può essere scritta come
\begin{equation}
 \xi\left(S,t+1\right)=\xi\left(S,t\right) \frac{eval\left(S,t\right)}{\bar{F}\left(t\right)}.
\label{Eq_CrescitaSchema2} 
\end{equation}
L'Equazione $\eqref{Eq_CrescitaSchema2}$ afferma che il numero di individui che presenta lo schema $S$ cresce ad ogni step temporale come il
rapporto del fitness di $S$ ed il fitness totale. Quindi possiamo affermare che uno schema con $eval\left(S,t\right)>\bar{F}\left(t\right)$
prolifererà nelle generazioni successive, mentre uno schema con fitness inferiore alla media tenderà ad estinguersi.
La formula afferma quindi che, in periodi di tempo lunghi, un buono schema prolifererà esponenzialmente mentre un cattivo schema si avvierà all'estinzione con velocità esponenziale.
\par A questo punto, è necessario analizzare i fenomeni di crossover e mutazione e di come essi impattino sulla possibilità che uno
schema prevalga o meno sugli altri col passare del tempo. \\
Un punto di crossover è selezionabile in $m-1$ punti possibili, dove $m$ è la lunghezza del vettore che rappresenta l'individuo. $\delta\left( S \right) / m-1$
rappresenta la probabilità che lo schema $S$ venga distrutto durante il corssover e quindi la probabilità di sopravvivenza $p_{cs}$ è
data da 
\begin{equation}
p_{cs}\ge1-p_c \frac{\delta \left(S\right)}{m-1}.
\nonumber
\end{equation}
Osserviamo che è presente il $\ge$ invece di $=$ perchè potrebbe capitare che lo schema $S$ sia spezzato e ricombinato esattamente.
Possiamo allora affermare che vale
\begin{equation}
\xi\left(S,t+1\right)\ge\xi\left(S,t\right) \frac{eval\left(S,t\right)}{\bar{F}\left(t\right)} \left( 1 -p_{c}\frac{\delta\left(S\right)}{m-1}\right),
\nonumber
\end{equation}
cioè il numero di individui alla generazione seguente che presentano lo schema $S$ è legato alla probabilità di sopravvivenza dello schema stesso.
Osserviamo che la formula afferma che schemi corti hanno una probabilità di sopravvivenza più elevata rispetto a schemi più lunghi.
\par Passiamo ora ad analizzare gli effetti della mutazione sul processo di evoluzione dello schema. Sia $p_m$ la probabilità
che un singolo componente dello schema cambi. Ogni mutazione è indipendente dalle altre così che la probabilità $p_{ms}\left(S\right)$ che uno schema sopravviva
alla mutazione è pari a
\begin{equation}
p_{ms}\left(S\right)=\left(1-p_m\right)^{\sigma\left(S\right)}.
\nonumber
\end{equation}
Si osservi quindi che più uno schema è lungo e minore sarà la sua proabiltà di sopravvivere alla mutazione. Assumendo che $p_m \ll 1$
otteniamo che $p_{ms}\left(S\right)\simeq 1-\sigma\left(S\right)$. Infine otteniamo che
\begin{equation}
\xi\left(S,t+1\right)\ge\xi\left(S,t\right) \frac{eval\left(S,t\right)}{\bar{F}\left(t\right)} \left( 1 -p_{cs}\frac{\delta\left(S\right)}{m-1}\right) \left(\left(1-p_m\right)^{\sigma\left(S)\right)}\right).
\label{schema_theorem}
\end{equation}

Arrivati a questo punto possiamo enunciare il teorema degli schemi.

\begin{thm}[Teorema Degli Schemi]
Con la nomenclatura introdotta sopra l'Equazione $\eqref{schema_theorem}$ vale.
\end{thm}
La formula afferma che quando si è in presenza di schemi corti con elevato valore della funzione di fitness e basso ordine ci sono maggiori
probabilità di sopravvivenza. Tali schemi nella seguente generazione dell'algoritmo genetico classico sono soggetti ad una crescita esponenziale
mentre gli altri schemi seguono una decrescita esponenziale.

\section{Viral Search}
In questa sezione introdurremo un nuovo tipo di algoritmo genetico che chiameremo \textit{Viral Search}. Tale algoritmo è un algoritmo di ottimizzazione che mira a trovare un minimo/massimo globale di una funzione data. Inizieremo a descrivere euristicamente il suo funzionamento per poi fornirne una descrizione più formale.

\subsection{L'idea}
L'idea sottesa all'algoritmo del \textit{Viral Search} è la seguente: si suppone che ci siano alcuni individui, che chiameremo virus, che si muovono aleatoriamente nello spazio dei parametri. Ognuno di essi in ogni punto in cui si trova \textquotedblleft valuta\textquotedblright\ le condizioni dello spazio circostante e decide di diffondersi se tale condizione è migliore della migliore condizione di diffusione verificatasi fino a quel momento. Se le condizioni sono favorevoli inizia a riprodursi continuamente, generando un' \textquotedblleft epidemia\textquotedblright\ più o meno localizzata al termine della quale si estingue, ricordando però quale è stata la condizione migliore che gli ha permesso di proliferare. Estinta l'epidemia il virus ricomincia a muoversi aleatoriamente alla ricerca di zone ulteriormente migliori dove poter generare una nuova pandemia.


\par In un'ottica di minimizzazione tramite algoritmi genetici il virus non è altro che il cromosoma, il mezzo di cui si serve il virus per valutare la possibilità di diffusione è la funzione obiettivo e la durata dell'epidemia è il numero di generazioni durante le quali si riproduce il virus. Al termine dell'esecuzione dell'algoritmo la migliore condizione di diffusione trovata dai virus che popolano lo spazio dei parametri altri non è se non un minimo della funzione obiettivo, possibilmente un minimo globale.
\par Ora che abbiamo illustrato euristicamente come opera il viral search, passiamo alla sua formalizzazione.

\subsection{L'algoritmo}
Iniziamo questa sezione col definire la terminologia che utilizzeremo in seguito. Siano
\begin{itemize}
 \item $N_g$: numero di generazioni, ovvero la durata del processo di diffusione del virus.
 \item $N_{gv}$: numero di generazioni virali, ovvero la durata di un'epidemia localizzata.
 \item $N_i$: numero di individui, cioè il numero di virus che esplorano lo spazio dei parametri.
 \item $N_{iv}$: numero di individui virali, cioè il numero di \textquotedblleft copie di se stesso\textquotedblright\ che il virus singolo genera quando trova una zona di diffusione favorevole.
 \item $Fobj$: valore funzione obiettivo.
 \item $FobjGlobal$: valore della funzione obiettivo globale.
 \item $BestIndividual$: migliore individuo trovato.
 \item $BestIndividualGlobal$: migliore individuo trovato globalmente.
\end{itemize}

L'algoritmo può essere sintetizzato come indicato nel seguente Algoritmo $\ref{Viral_Search_Algorithm_listato}$.

\begin{algorithm}
\caption{Viral Search}
\label{Viral_Search_Algorithm_listato}
\begin{algorithmic}[1]
\STATE Set $N_g$,$N_{gv}$,$N_i$,$N_{iv}$,$Fobj$,$FobjGlobal$ and $t=1$.
\STATE Generate new population $P\left(t\right)$. \label{Vs:is2}
\WHILE{$t\le Ng$} 
\FOR{$i=1$ \TO $i=Ni$}
\STATE Evaluate Objective Function and set it to value $Fobj_i$.
\IF{$Fobj_i\le FobjGlobal$}
\STATE Start viral diffusion and set $BestIndividual_i$ and $Fobj_i$ as the best local indivual found and its associated value of objective function.
\ENDIF
\IF{$Fobj_i\le FobjGlobal$}
\STATE Set $FobjGlobal=Fobj_i$ 
\STATE Set $BestIndividualGlobal=BestIndividual_i$
\ENDIF
\ENDFOR
\STATE Move randomly viral population $P\left(t\right)$.
\ENDWHILE
\end{algorithmic}
\end{algorithm}

Analizziamo i singoli passi dell'algoritmo.
\par La prima istruzione è la fase di inizializzazione dei parametri. Si sceglie il numero di generazioni $N_g$ ed il numero di generazioni virali $N_{gv}$. 
Maggiore è $N_g$ e maggiore sarà la durata dell'esplorazione globale dello spazio dei parametri. $N_{gv}$ invece determina la durata della ricerca del minimo locale in una certa area selezionata. Se si vuole privilegiare una ricerca di tipo globale, sospettando l'esistenza di più minimi locali,
potrebbe essere opportuno scegliere $N_g\gg N_{gv}$.\\
Altra scelta da compiere è il numero di individui virali che si muovono alla ricerca di zone di diffusione migliori (parametro $N_i$) ed il numero di
copie virali generate ogni volta che si presenta una zona di diffusione migliore delle precedenti (parametro $N_{iv}$). Come nel caso precedente, se
si volesse privilegiare una ricerca di tipo globale è opportuno scegliere $N_i\gg N_{iv}$. Un buon accorgimento è scegliere il numero di individui $N_i=N_i\left(N_p\right)$ come
funzione crescente del numero di parametri $\left(N_p\right)$ così che possa essere garantito un numero sufficientemente alto di individui in grado di esplorare adeguatamente
tutto lo spazio dei parametri. Vanno anche assegnati i valori iniziali a $Fobj$ ed a $FobjGlobal$ che, inizialmente, possono essere posti pari a $+ \infty$, mentre i valori di $BestIndividual$ e di
$BestIndividualGlobal$ risulteranno non assegnati.
\par L'istruzione ~\ref{Vs:is2} è dedicata alla generazione della popolazione iniziale di individui. Tale popolazione può essere generata in vari modi, avendo
però cura di rispettare i vincoli sui parametri. Supponiamo che $\boldsymbol{ub}$ sia il vettore dei limiti superiori dei parametri e $\boldsymbol{lb}$
quello dei limiti inferiori. La popolazione viene generata tenendo conto di questi limiti, creando una nuvola di $N_i$ punti casuali oppure cercando di 
posizionare gli individui di modo che coprano il più uniformemente possibile lo spazio dei parametri. 
\par Dal passo successivo inizia il vero e proprio algoritmo. Finchè non viene raggiunto il limite di generazione $N_g$, oppure, in alternativa, viene soddisfatto 
un qualche criterio basato sulla funzione obiettivo \cite{Manicardi2013}, per ogni invidividuo $i$ in $N_i$ viene valutata la funzione obiettivo. Se il valore della funzione obiettivo è minore del valore della funzione obiettivo globale $FobjGlobal$ viene generata una nuvola di punti localizzata e viene eseguito un algoritmo di ricerca locale di tipo genetico. In questo
articolo per tutte le ricerche locali verrà utilizzato il \textit{Differential Evolution}, la cui descrizione può essere trovata in \cite{Storn1997}. L'area in cui viene effettuata
la ricerca può essere scelta in vari modi: un modo è selezionare l'individuo che ha identificato un miglioramento per la funzione obiettivo globale e selezionarlo
come centro di cubo $N_p$-dimensionale in cui avverrà la \textquotedblleft diffusione\textquotedblright\. Terminata la ricerca locale il valore 
della funzione obiettivo globale $FobjGlobal$ diventa il valore della funzione obiettivo associata all'individuo migliore dell'intera popolazione $BestIndividualGlobal$ cioè il cromosoma che ha realizzato il minimo
della funzione obiettivo globale. A questo punto la popolazione $P\left(t\right)$ viene mossa aleatoriamente o pseudo-casualmente e l'algoritmo può ripetersi fino a $N_g$.
\par Il movimento della popolazione deve essere oggetto di un'attenta analisi. Oltre ad essere possibile che un individuo esca dallo spazio dei parametri, problematica
facilmente risolvibile, potrebbe capitare che, soprattutto in spazi particolarmente grandi, un movimento casuale non riesca a riempire completamente
una particolare zona. Per ovviare a questo problema una strategia potrebbe essere la seguente: definiamo un numero $N_c$ di 
\textquotedblleft centri \textquotedblright, ovvero dei punti fissi nello spazione $N_p$-dimensionale dei parametri. Per ogni individuo $i$, ad ogni generazione, si
calcola a quale centro esso è più vicino. L'invidividuo $i$ si muove aleatoriamente per un certo numero di generazioni, dopo di che si calcola quali centri siano stati meno visitati degli altri e si muovono alcuni individui dai centri più visitati ai centri meno visitati. Questo procedimento garantisce un'esplorazione
più uniforme dello spazio dei parametri soprattutto nel caso in cui esso diventi particolarmente ampio, ovvero nei casi in cui 
$\left| \boldsymbol{ub}_j-\boldsymbol{lb}_j \right|$ sia elevato per qualche $j \in 1,\dots,N_p$. Nel caso in cui si vogliano usare dei \textquotedblleft centri \textquotedblright è necessario
garantire una distribuzione uniforme dei centri, oltre ad utilizzarne un numero adeguato alla dimensione del problema, intesa sia come numero di parametri coinvolti nell'ottimizzazione sia come dimensione dell' area di esplorazione. Nel caso in cui si avessero delle informazioni a priori sulla funzione da minimizzare è possibile disporre i centri in modo tale che vengano posti nei punti in cui si sospetta si possa trovare il minimo cercato.
\par Da ultimo osserviamo che questo algoritmo, come tutti gli algoritmi genetici, è intrinsecamente parallelizzabile. La parallelizzazione può avvenire facilmente dividendo l'insieme $\Omega$ su cui si vuole minimizzare la funzione in $m$ insiemi $A_{i}$ tali che $\bigcup_1^{i} m A_{i} = \Omega$ ed affidando la ricerca su ognuno dei sottoinsiemi $A_{i}$ ad un diverso calcolatore.

\section{Risultati Numerici}
In questa sezione presenteremo i risultati numerici che testimoniano la bontà dell'algoritmo presentato in precedenza. L'algoritmo verrà testato per ricercare i punti di massimo o minimo di funzioni usate tipicamente per testare l'efficacia di algoritmi di minimizzazione. Queste funzioni sono difficili da ottimizzare perchè presentano molti minimi locali, hanno zone a derivata nulla, hanno punti estremanti variabili nel tempo oppure non sono differenziabili. Una lista di queste funzioni può essere trovata in \cite{WikiOptListFun2015}. Analizzeremo il comportamento dell'algoritmo e ne ossereveremo il costo computazionale. Tutti gli esprerimenti sono stati condotti sul un PC desktop AMD Athlon(tm) II X4 620 Processor 2.6 GHz e 4 GB di RAM.

\subsection{La funzione di Rosenbrock}
\begin{figure}[!h]
\centering
\includegraphics[scale=0.5]{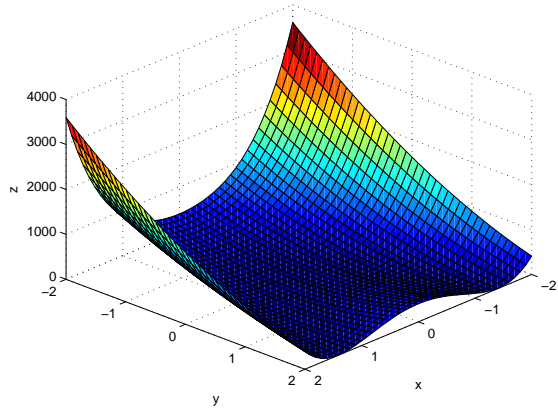}
\caption{Funzione di Rosenbrock}
\label{fig_funzione_rosenbrock}
\end{figure}

La funzione di Rosenbrock è una funzione non convessa largamente utilizzata per testare gli algoritmi di minimizzazione. La funzione di Rosenbrock in due dimensioni ha la seguente espressione analitica:
\begin{equation}
f\left(x,y\right)=\left(a-x \right)^2 + b\left(y-x^2 \right)^2
\nonumber
\end{equation}
dove solitamente $a=1$ e $b=100$. Essa presenta una \textquotedblleft valle\textquotedblright\ di minimi ed un unico punto di minimo globale posto in $\left(a,a^2\right)$. Trovare la valle è generalmente semplice, ma convergere al minimo globale è complesso. La funzione di Rosenbrock, nel suo caso bidimensionale, è mostrata in Figura $\eqref{fig_funzione_rosenbrock}$.

In questo esperimento numerico fissiamo come estremi del dominio il quadrato di vertici $\left(-3,-3\right)$, $\left(-3,3\right)$,$\left(3,-3\right)$, $\left(3,3\right)$. Poniamo
$N_p=2$ ed, inizialmente, $N_i=20*N_p$, $N_g=1000$, $N_{iv}=75*N_p$ ed $N_{gv}=75$. Utilizziamo, inoltre, un metodo di movimento degli individui basato sulla definizione dei centri, così da cercare di esplorare tutte le zone in maniera uniforme, ponendo $N_c=7$ dove $N_c$ è il numero di centri per asse. In totale si avranno, essendo lo spazio di ricerca bidimensionale, $N_{c}^{2}$ centri. Da questa osservazione deduciamo che definire in tal modo i centri costituisce una criticità al crescere del numero delle dimensioni siccome la crescita del numero di centri è esponenziale nel numero dei parametri da cui dipende la funzione obiettivo. A questo problema si può ovviare utilizzando meno centri, parallelizzando l'algoritmo, in modo tale da analizzare zone più piccole in parallelo oppure si può optare per un movimento completamente casuale degli individui. Per quanto rigarda la funzione di Rosenbrock scegliamo $a=1$ e $b=100$, così da avere un minimo in $\left( 1,1 \right)$. 
La correttezza dell'algoritmo si evince dai risultati mostrati in Tabella $\eqref{Tabella_Rosenbrock_001}$.

\begin{table}[!ht]
\centering
\begin{tabular}{*{2}{c}}
\toprule
\multicolumn{2}{c}{Viral Search Rosenbrock} \\
\midrule
 Minimo Analitico & Minimo Numerico\\
 \midrule
 $\left(1,1,0\right)$ & $\left(1.01, 1.04, 3.9436e-04\right)$\\
 \bottomrule
\end{tabular}
\caption{Test numerico sulla funzione di Rosenbrok.}
\label{Tabella_Rosenbrock_001}
\end{table}
 
 E' possibile aumentare la precisione dell'algoritmo aumentando o il numero di generazioni $N_g$ oppure il numero di individui $N_i$. In Tabella $\eqref{Tabella_Rosenbrock_002}$ sono riportati i risultati ottenuti aumentando il numero di individui $N_i$ ed il numero di generazioni $N_g$. Notiamo che l'algoritmo, a discapito della velocità, aumenta la sua precisione.
 
 \begin{table}[!ht]
\centering
\begin{tabular}{*{6}{c}}
\toprule
$N_i$ & $N_g$ & $x$ & $y$ &  $z$ & $T\left(s\right)$\\
\midrule
5 & 50 & 1.577528 & 2.499517 & 0.345470 & 2.327730 \\ 
10 & 75 & 1.188800 & 1.413805 & 0.035677 & 1.041527 \\ 
30 & 100 & 1.063673 & 1.131522 & 0.004056 & 0.957370 \\ 
60 & 200 & 1.000619 & 1.001172 & 0.000001 & 1.336336 \\ 
100 & 300 & 0.956890 & 0.915136 & 0.001884 & 3.798816 \\ 
400 & 1200 & 1.000222 & 1.000432 & 0.000000 & 50.314480 \\ 
 \bottomrule
\end{tabular}
\caption{Convergenza dell'algoritmo all'aumentare del numero di individui e di generazioni. $N_{iv}=150$ e $N_{gv}=75$}
\label{Tabella_Rosenbrock_002}
\end{table}

\begin{figure}[!ht]
\centering
\includegraphics[scale=0.5]{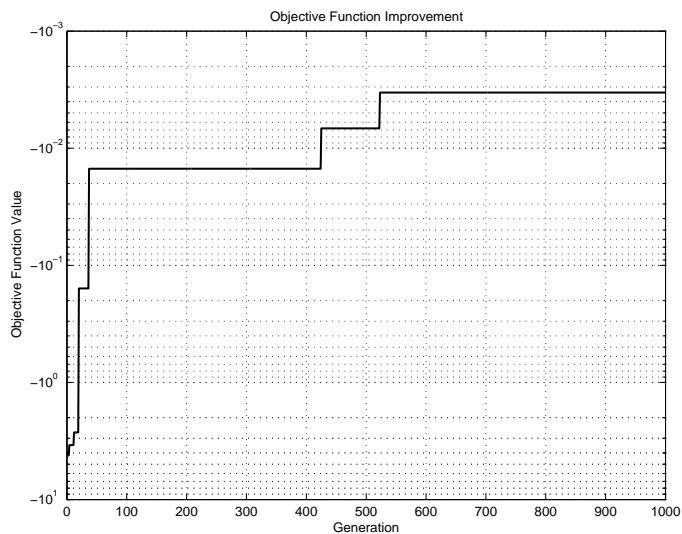}
\caption{Miglioramento della funzione obiettivo all'aumentare del numero di generazioni con $N_i=40$.}
\label{fig_miglioramento_algoritmi_Ng_Rosenbrock}
\end{figure}

Un'idea di con che velocità migliori l'algoritmo all'aumentare del numero di generazioni a parità di individui $N_i$ è invece data dalla Figura $\eqref{fig_miglioramento_algoritmi_Ng_Rosenbrock}$: sull'asse delle $x$ è posto il numero di generazioni mentre sull'asse delle $y$ è mostrato il valore della funzione obiettivo. Si nota chiaramente che all'aumentare del numero di generazioni il valore della funzione obiettivo diventa prossimo allo $0$ a supporto della dimostrazione di convergenza dell'algoritmo.

Ciò che invece non incide eccessivamente sulla velocità di convergenza dell'algoritmo è il numero di generazioni virali $N_{gv}$ ed il numero di individui virali, cioè $N_{iv}$. Ogni volta che viene chiamato l'algoritmo su una determinata porzione di area è opportuno usare un numero adeguato di $N_{gv}$ e $N_{iv},$, dipendenti dalla dimensione dell'area da esplorare, tale da garantire la corvergenza dell'algoritmo genetico locale. Questo fatto è ben visibile dai risultati riportati in Tabella $\eqref{Tabella_Rosenbrock_003}$: si vede che non si nota nessun rilevante miglioramento a partire da $N_{iv}=30$ e $N_{gv}=100$, segno che probabilmente una scelta ottimale per i parametri $N_{iv}$ e $N_{gv}$ sia proprio quest'ultima. Osserviamo che tanto più grande è l'area da esplorare tanto maggiori dovranno essere i parametri $N_{iv}$ e $N_{gv}$ per poter ottenere buoni risultati. 

 \begin{table}[!ht]
\centering
\begin{tabular}{*{6}{c}}
\toprule
$N_{iv}$ & $N_{gv}$ & $x$ & $y$ &  $z$ & $T\left(s\right)$\\
\midrule
5 & 50 & 0.543221 & 0.294429 & 0.208690 & 0.517534 \\ 
10 & 75 & 1.003317 & 1.006599 & 0.000011 & 0.555820 \\ 
30 & 100 & 0.856556 & 0.733349 & 0.020588 & 0.687655 \\ 
60 & 200 & 1.084461 & 1.176444 & 0.007149 & 1.448558 \\ 
100 & 300 & 0.973215 & 0.947145 & 0.000717 & 2.077930 \\ 
400 & 500 & 0.999746 & 0.999457 & 0.000000 & 17.368053 \\ 
400 & 1200 & 1.000222 & 1.000432 & 0.000000 & 50.314480 \\ 
 \bottomrule
\end{tabular}
\caption{Convergenza dell'algoritmo all'aumentare del numero di individui virali e di generazioni virali. $N_{i}=40$ e $N_{g}=100$}
\label{Tabella_Rosenbrock_003}
\end{table}

\subsection{La funzione di Schaffer}

\par Consideriamo ora un'altra funzione che viene tipicamente utilizzata nei test di ottimizzazione: la funzione di Schaffer. Essa è una funzione che ammette la seguente espressione analitica:

\begin{equation}
f\left(x,y\right)= 0.5 +\frac{sin^2\left(x^2-y^2\right)-0.5}{1+0.001\left(x^2+y^2\right)}
\nonumber
\end{equation}

\begin{figure}[!ht]
\centering
\includegraphics[scale=0.5]{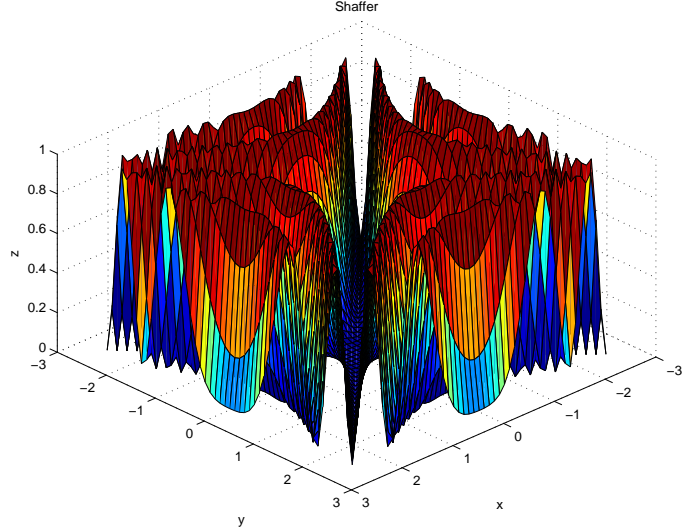}
\caption{Funzione di Schaffer}
\label{fig_funzione_schaffer}
\end{figure}

Tale funzione è di tipo oscillatorio, presenta numerosi limiti locali, un solo minimo globale nel punto $\left(0,0\right)$ ed è rappresentata in Figura $\eqref{fig_funzione_schaffer}$. L'elevato numero di minimi locali rende molto difficile individuare l'unico minimo globale.

Proviamo ora a cercare il minimo assoluto utilizzando l'algoritmo di Viral Search. Ripetendo un esperimento analogo a quello eseguito per la funzione di Rosenbrock, sempre sul rettangolo di vertici  $\left(-3,-3\right)$, $\left(-3,3\right)$,$\left(3,-3\right)$, $\left(3,3\right)$, otteniamo i risultati riportati in Tabella $\eqref{Tabella_Schaffer_001}$. In questo caso osserviamo che è necessario un maggior numero di individui $N_i$ per da garantire un buon ricoprimento dello spazio delle soluzioni. Man mano che lo spazio delle soluzioni diventa ampio, sia in numero di variabili, sia in estensione si verifica un aumento del tempo computazionale ed un aumento del numero di individui necessari per poter garantire la covergenza. In tal caso, un'appropriata scelta dei centri potrebbe risultare vincente, ad esempio posizionandoli nelle zone in cui ci si aspetta che la funzione possa presentare dei minimi.

 \begin{table}[!ht]
\centering
\begin{tabular}{*{6}{c}}
\toprule
$N_i$ & $N_g$ & $x$ & $y$ &  $z$ & $T\left(s\right)$\\
\midrule
10 & 50 & 3.877474 & 7.042075 & 0.058862 & 1.703371 \\ 
50 & 75 & -1.808829 & -7.528476 & 0.054960 & 2.266957 \\ 
100 & 100 & -0.973998 & -2.689123 & 0.008081 & 2.597825 \\ 
400 & 150 & -0.096062 & 0.098659 & 0.000019 & 6.383318 \\ 
1000 & 200 & 0.470023 & -0.470468 & 0.000442 & 21.300443 \\ 
1500 & 300 & -0.205643 & 0.205748 & 0.000085 & 46.546393 \\ 
 \bottomrule
\end{tabular}
\caption{Convergenza dell'algoritmo all'aumentare del numero di individui e di generazioni. La convergenza non monotona decrescente al minimo è 
dovuta alla diversa inizializzazione del seme per la generazione dei numeri aleatori necessari all'algoritmo. $N_{iv}=150$ e $N_{gv}=75$}
\label{Tabella_Schaffer_001}
\end{table}

Dai risultati si nota come anche nel caso in cui si debba cercare il minimo globale di una funzione con molti minimi locali l'algoritmo di Viral Search riesce, in un tempo ragionevole, a trovare il minimo globale, cosa impossibile per un metodo basato sul semplice calcolo del gradiente quale può essere il metodo del gradiente coniugato. Ancora una volta, come nel caso precednte, notiamo che  aumentando i valori di $N_{iv}$ e $N_{gv}$ è possibile migliorare ulteriormente il risultato ottenuto a discapito, però, della velocità di esecuzione.

\subsection{Una funzione tempo variante}
\begin{figure}[!ht]
 \centering
 \begin{subfigure}[Superificie per $t=1$.]
  {
  \includegraphics[scale=0.32]{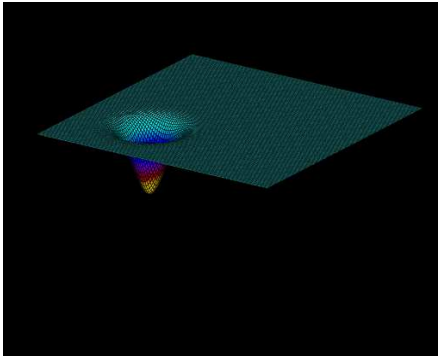}
  }
 \end{subfigure}
  \begin{subfigure}[Superificie per $t=30000$.]
  {
  \includegraphics[scale=0.32]{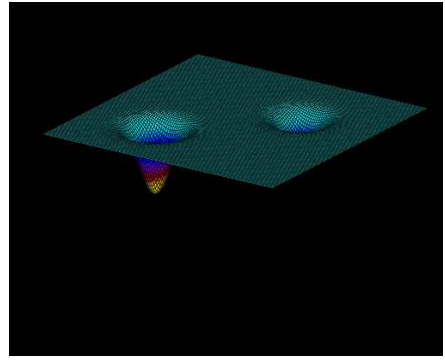}
  }
 \end{subfigure}
  \begin{subfigure}[Superificie per $t=60000$.]
  {
  \includegraphics[scale=0.32]{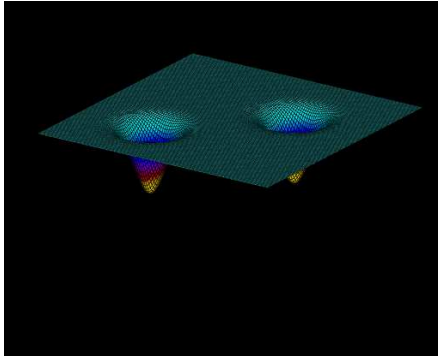}
  }
 \end{subfigure}
  \begin{subfigure}[Superificie per $t=90000$.]
  {
  \includegraphics[scale=0.32]{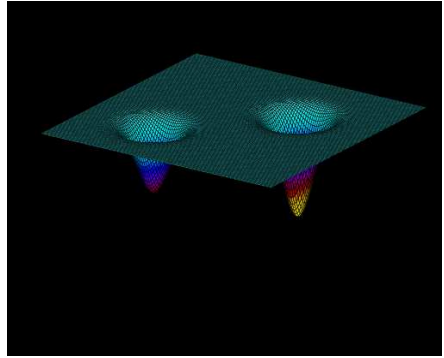}
  }
 \end{subfigure}
 \caption{Superficie tempo variante. In $t=1$ è presente l'unico minimo globale in $\left(-\frac{5}{2} -\frac{5}{2} \right)$, mentre
 in $t=9000$ il minimo globale si trova in $\left(2,2\right)$.}
 \label{fig_funzione_TempoVariante}
\end{figure}

Mostriamo ora come un algoritmo di questo tipo possa essere utilizzato nel caso in cui si voglia minimizzare una funzione tempo variante. 
Supponiamo di voler minimizzare la seguente funzione, definita per $x\in\left[-6,6\right]$ e $y\in\left[-6,6\right]$ . 

\begin{equation}
f\left(t,x,y\right)= -2e^{\left(\left(x+\frac{5}{2}\right)^2+\left(y+\frac{5}{2}\right)^2\right)} -\frac{t}{1000}e^{\left( \left(x-2\right)^2+\left(y-2\right)^2\right)}
\nonumber
\end{equation}
Tale funzione presenta inizialmente un minimo globale in $\left(-\frac{5}{2} -\frac{5}{2}\right)$ ma, con lo scorrere del tempo, appare un minimo globale in $\left(2,2\right)$, come visibile in Figura $\eqref{fig_funzione_TempoVariante}$.

In Tabella $\eqref{Tabella_funTimeVariant_001}$ osserviamo come l'algoritmo identifichi inizialmente il minimo in $\left(-\frac{5}{2} -\frac{5}{2} \right)$,
dopo di che, al mutare del minimo, esso trovi il nuovo minimo in $\left(2,2\right)$. Questa potenzialità dell'algoritmo può essere sfruttata nei problemi di ottimizzazione tempo varianti in cui la funzione cambia nel tempo. Inoltre, scegliendo un numero non troppo elevato di individui $N_{i}$ è possibile mantenere attivo l'algoritmo in background utilizzando, di fatto, una minima potenza di calcolo. Solo nel momento in cui verrà trovato un nuovo minimo verrà richiesta la potenza di calcolo necessaria per poter esplorare la zona di interesse.
 \begin{table}[!ht]
\centering
\begin{tabular}{*{4}{c}}
\toprule
$t$ & $x$ & $y$ & $z$ \\
\midrule
10000 & -2.505741 & -2.501649 & -1.999929\\
20000 & -2.504568 & -2.496689 & -1.999936\\
30000 & -2.504568 & -2.496689 & -1.999936\\
40000 & -2.499187 & -2.501268 & -1.999995\\
50000 & -2.499187 & -2.501268 & -1.999995\\
60000 & -2.499187 & -2.501268 & -1.999995\\
70000 & 1.957308 & 1.967954 & -2.325962\\
80000 & 2.007791 & 1.929722 & -2.649089\\
90000 & 1.985048 & 2.006218 & -2.998347\\
 \bottomrule
\end{tabular}
\caption{Minimo globale trovato dall'algoritmo con parametri: $N_i=100$, $N_g=90000$, $N_{iv}=150$, $N_{gv}=100$.}
\label{Tabella_funTimeVariant_001}
\end{table}

\subsection{Una funzione multidimensionale}
\par
\begin{figure}
	\centering
	\includegraphics[scale=0.5]{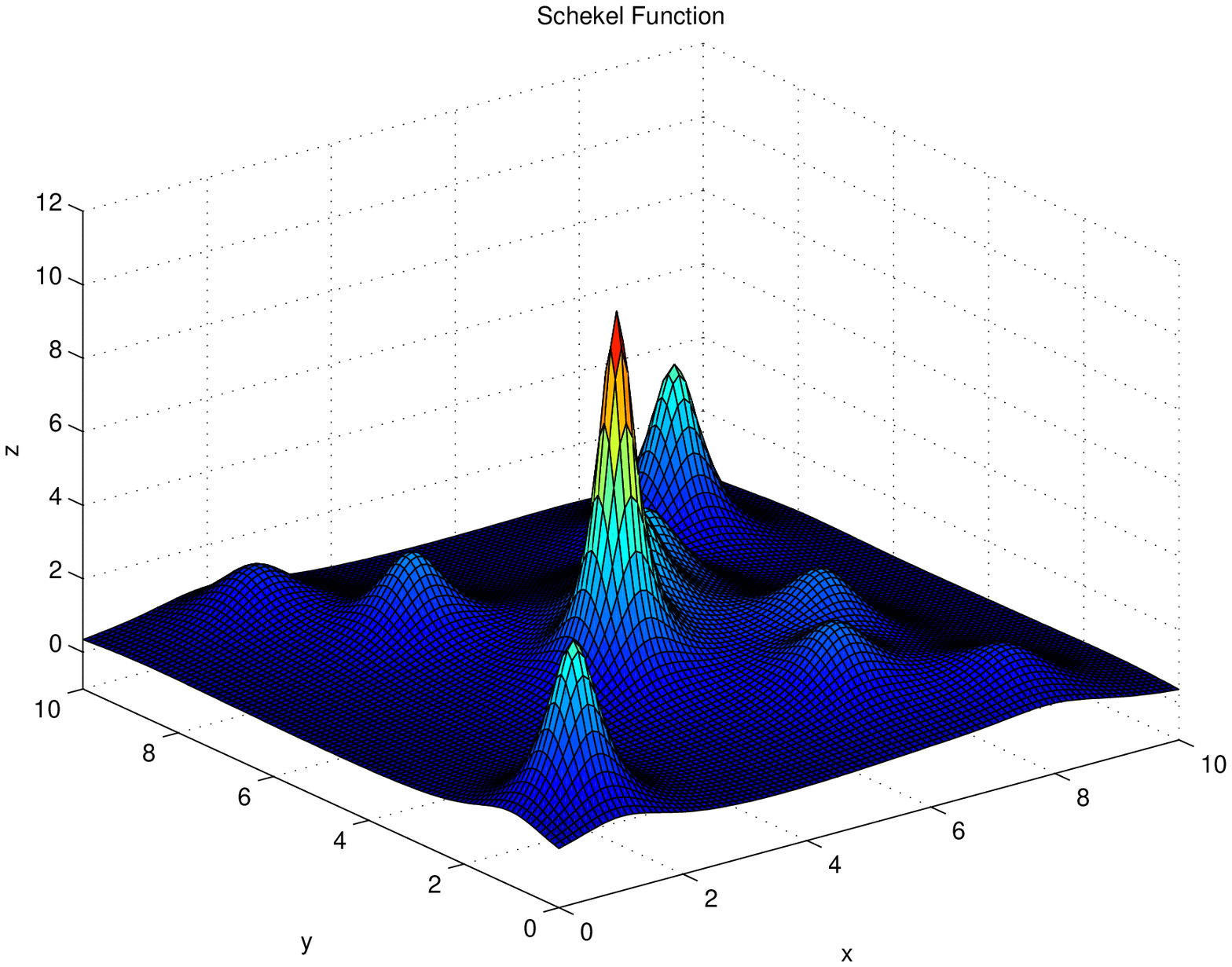}
	\caption{Funzione di Schekel tridimensionale, presenta un massimo globale in $x^{\sharp}=\left(4,4\right)$.}
	\label{fig_Schekel_2D_Plot}
\end{figure}

\begin{figure}
	\centering
	\includegraphics[scale=0.5]{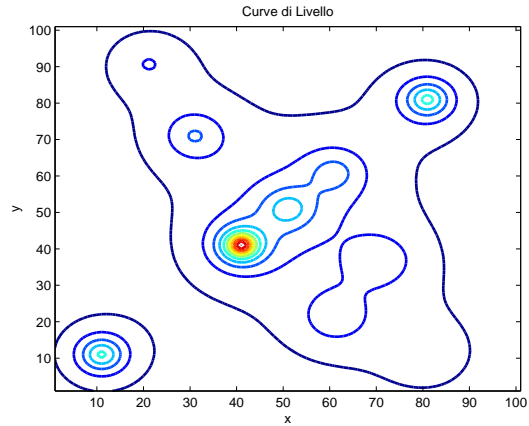}
	\caption{Curve di livello per la funzione di Schekel.}
	\label{fig_Schekel_2D_Curves}
\end{figure}

 Da ultimo proponiamo una massimizzazione di una funzione multidimensionale detta Funzione di Shekel, Essa nel caso $n$-dimensionale con $m$ minimi presenta la forma seguente:
\begin{equation}
f\left(\vec{x}\right)=- \sum_{i=1}^{m} \left(c_i \sum_{j=1}^{n} \left(x_j-a_{ji}\right)^2 \right)^{-1}.
\end{equation}
Nell'esperimento che condurremo consideriamo 
\begin{equation}
\boldsymbol{c}=\frac{1}{10}  \left\{1,2,2,4,4,6,3,7,5,5 \right\}
\nonumber
\end{equation}
e
\begin{equation}
A=\begin{bmatrix}
4 & 1 & 8 & 6 & 3 & 2 & 5 & 8 & 6 & 7 \\
4 &  1 & 8 & 6 & 7 & 9 & 5 & 1 & 2 & 3.6 \\
4 & 1 & 8 & 6 & 3 & 2 & 5 & 8 & 6 & 7 \\
4 & 1 & 8 & 6 & 7 & 9 & 3 & 1 & 2 & 3\\
\end{bmatrix}
\end{equation}
Sotto tali condizioni, nella regione $0\le x_i\le 10$, è presente un massimo globale $x^{\sharp}=\left(4,4,4,4\right)$, e $f\left(x^{\sharp}\right)=10.692$. 
Per dare un'idea di come si presenta tale funzione ne riportiamo il caso bidimensionale in Figura $\eqref{fig_Schekel_2D_Plot}$ e le relative curve di livello in Figura $\eqref{fig_Schekel_2D_Curves}$.

Lanciamo l'algoritmo proposto precedentemente considerando diversi valori di $N_i$ e $N_g$. Osserviamo dai risultati in Tabella $\eqref{Tabella_Schakel_005}$ che all'aumentare del numero di generazioni e di individui l'algoritmo arriva a convergenza, individuando il massimo globale $x^{\sharp}$. 

\begin{table}[!ht]
	\centering
	\begin{tabular}{*{8}{c}}
		\toprule
		$N_{i}$ & $N_{g}$ & $x\left(1\right)$ & $x\left(2\right)$ & $x\left(3\right)$& $x\left(4\right)$ &  $Val$ & $T\left(s\right)$\\
		\midrule
		5 & 50 & 6.2702 & 1.6316 & 6.1299 & 2.0350 & 1.8386 & 0.0908 \\ 
		10 & 75 & 3.9449 & 3.3552 & 3.6989 & 4.0596 & 2.0552 & 0.0893 \\ 
		30 & 100 & 3.9745 & 4.0870 & 4.0552 & 4.3697 & 4.5422 & 0.0782 \\ 
		60 & 200 & 3.8479 & 4.2399 & 3.7180 & 3.7828 & 3.7622 & 0.2350 \\ 
		100 & 300 & 3.6842 & 4.1257 & 3.9470 & 4.4123 & 3.0385& 0.2931 \\ 
		400 & 1200 & 3.9609 & 3.9791 & 3.9159 & 3.9956 & 9.6834 & 4.0686 \\ 
		800 & 1500 & 3.9977 & 3.9588 & 3.9570 & 4.0526 & 9.9235 & 9.9994 \\ 
		1000 & 2000 & 4.0051 & 3.9984 & 4.0108 & 4.0576 & 10.1953 & 16.6048 \\ 
		2000 & 3000 & 3.9706 & 3.8976 & 3.9277 & 4.0657 & 8.7743 & 49.7099 \\ 
		5000 & 5000 & 4.0124 & 3.9811 & 3.9981 & 3.9879 & 10.4688 & 206.6701 \\
		\bottomrule
	\end{tabular}
	\caption{Convergenza dell'algoritmo all'aumentare del numero di individui e di generazioni. $N_{iv}=300$ e $N_{gv}=75$}
	\label{Tabella_Schakel_005}
\end{table}

Ovviamente  all'aumentare del numero di dimensioni corrisponde un aumento del tempo di computazione necessario per arrivare a convergenza. Ciò è dovuto, come osservato più volte, al fatto che, per garantire un buon risultato di ottimizzazione, è necessario utilizzare un sempre maggior numero di individui $N_i$ ed $N_g$ a scapito della velocità di computazione.

\clearpage

\section{Conclusioni}
Si è visto come l'algorimo di Viral Search riesca a trovare i punti estremanti di funzioni di difficile ottimizzazione grazie ad un approccio di tipo euristico. 
Gli sviluppi possibili di questo algoritmo sono molti e tanti sono i miglioramenti implementabili a partire dalla sua parallelizzazione. Altro miglioramento implementabile è cercare di ridurre l'aleatorietà del movimento degli individui nello spazio delle soluzioni. E' noto e facilmente verificabile numericamente che all'aumentare del numero delle dimensioni il movimento
totalmente casuale non riesce a coprire adeguatamente lo spazio delle soluzioni a meno di aumentare spropositamente il numero di individui $N_i$. Un'idea potrebbe essere quella di far muovere gli individui per inerzia verso le zone di maggior incremento della funzione, seguendo quello che è l'approccio dei metodi \textit{steepest-descend}, mantenendo comunque una componente aleatoria nel movimento così da evitare il ristagno in minimi locali. \\
Uno dei punti forza dell'algoritmo è la gestione dell'allocazione di memoria. Infatti tale algoritmo esegue una notevole allocazione di memoria solo quando viene lanciato l'algoritmo di ricerca locale. La memoria RAM resta occupata solo per il tempo necessario all'esecuzione dell'algoritmo di ricerca locale, dopo di che essa viene liberata e torna disponibile per altre applicazioni e calcoli. Ciò permette all'algoritmo essere eseguito con un basso dispedio di risorse e di richiedere maggiore potenza di calcolo solo nel momento in cui essa è effettivamente necessaria.

\bibliographystyle{plain}
\bibliography{Bibliografia}

\begin{thebibliography}{10}

\bibitem{Bridges1987}
Claiton~L. Bridges and David~E. Goldberg.
\newblock 2nd int'l conf. n genetic algorithms and their applications.
\newblock In {\em An analysis of reproduction and crossover in a binary-coded
  genetic algorithm}, 1987.

\bibitem{Dantzig1963}
G.B. Dantzig.
\newblock {\em Linear programming and extensions}.
\newblock Princeton University Press, 1963.

\bibitem{Dantzig2003}
G.B. Dantzig.
\newblock {\em Linear programming 2: Theory and extensions}.
\newblock Springer-Verlag, 2003.

\bibitem{Fogel1995}
D.B.. Fogel.
\newblock {\em Evolutionary computation: toward a new philosophy of machine
  intelligence}.
\newblock IEEE Press, New York, 1995.

\bibitem{Hestens1952}
M.R. Hestens and E.~Stiefel.
\newblock Methods of conjugate gradients for solving linear systems.
\newblock {\em Journal of Research of the National Bureau of Standards}, 6:49,
  1952.

\bibitem{Holland1975}
John~H. Holland.
\newblock {\em Adaption in Natural and Artificial Systems}.
\newblock Bradford Books, 1975.

\bibitem{Koza1992}
J.R. Koza.
\newblock {\em Genetic programming: on the programming of computers by means of
  natural selection}.
\newblock MIT Press, Cambridge, 1992.

\bibitem{Manicardi2013}
A.~Manicardi.
\newblock Algoritmi genetici a supporto del problema della calibrazione,
  2012/2013.

\bibitem{Poli2008}
R.~Poli, W.~B. Langdon, and N.~F. McPhee.
\newblock A field guide to genetic programming.
\newblock \url{lulu.com}, 2008.

\bibitem{Rachemberg1971}
I.~Rachemberg.
\newblock {\em Evolutions strategie - Optimierung technischer Systeme nach
  Prinzipien der biologischen Evolution}.
\newblock PhD thesis, Technical University, Berlin, 1971.

\bibitem{Storn1997}
R.~Storn and K.~Price.
\newblock Differential evolution - a simple and efficienti heuristic for global
  optimization over continuo spaces.
\newblock {\em Journal of Global Optimization}, 11:519--523, 1997.

\bibitem{WikiOptListFun2015}
Wikipedia.
\newblock Test function for optimization.
\newblock \url{https://en.wikipedia.org/wiki/Test_functions_for_optimization},
  2015.

\bibitem{Wright1991}
Alden~H. Wright.
\newblock Genetic algorithms for real parameter optimization, 1991.

\end{thebibliography}

\end{document}